\def\year{2019}\relax
\documentclass[letterpaper]{article} %DO NOT CHANGE THIS
\usepackage{aaai19}  %Required
\usepackage{times}  %Required
\usepackage{helvet}  %Required
\usepackage{courier}  %Required
\usepackage{url}  %Required
\usepackage{graphicx}  %Required

\usepackage{multirow}
\usepackage{amsmath}
\usepackage{amsfonts}
\usepackage{times}
\usepackage{graphics}
\usepackage[british]{babel}
\usepackage{longtable}
\usepackage{rotating}
\usepackage{enumitem}
\usepackage{sistyle}

\usepackage{algorithm}
\usepackage[noend]{algpseudocode}

\begin{document}
\title{HAS-QA: Hierarchical Answer Spans Model \\
		for Open-domain Question Answering}
\author{Liang Pang${}^{\dag}$, Yanyan Lan${}^{\dag*}$, Jiafeng Guo${}^{\dag}$, Jun Xu${}^{\dag}$, Lixin Su${}^{\dag}$ \and Xueqi Cheng${}^{\dag}$\\
$\dag$CAS Key Laboratory of Network Data Science and Technology, \\ Institute of Computing Technology, Chinese Academy of Sciences, Beijing, China\\
$\dag$University of Chinese Academy of Sciences, Beijing, China\\
$*$Department of Statistics, University of California, Berkeley\\
\{pangliang,lanyanyan,guojiafeng,sulixin,cxq\}@ict.ac.cn, 
junxu@ruc.edu.cn\\
}

\maketitle
\begin{abstract}
This paper is concerned with open-domain question answering (i.e., OpenQA). Recently, some works have viewed this problem as a reading comprehension (RC) task, and directly applied successful RC models to it. However, the performances of such models are not so good as that in the RC task. In our opinion, the perspective of RC ignores three characteristics in OpenQA task: 
1) many paragraphs without the answer span are included in the data collection; 
2) multiple answer spans may exist within one given paragraph;
3) the end position of an answer span is dependent with the start position.
In this paper, we first propose a new probabilistic formulation of OpenQA, based on a three-level hierarchical structure, i.e.,~the question level, the paragraph level and the answer span level. Then a Hierarchical Answer Spans Model (HAS-QA) is designed to capture each probability.
HAS-QA has the ability to tackle the above three problems, and experiments on public OpenQA datasets show that it significantly outperforms traditional RC baselines and recent OpenQA baselines. 
\end{abstract}

\section{Introduction}
\label{sec:introduction}

Open-domain question answering (OpenQA) aims to seek answers for a broad range of questions from a large knowledge sources, e.g.,~structured knowledge bases~\cite{berant2013semantic,mou2017coupling} and unstructured documents from search engine~\cite{ferrucci2010building}. 
In this paper we focus on the OpenQA task with the unstructured knowledge sources retrieved by search engine. 

\begin{figure}
  \centering
  \includegraphics[width=0.45\textwidth]{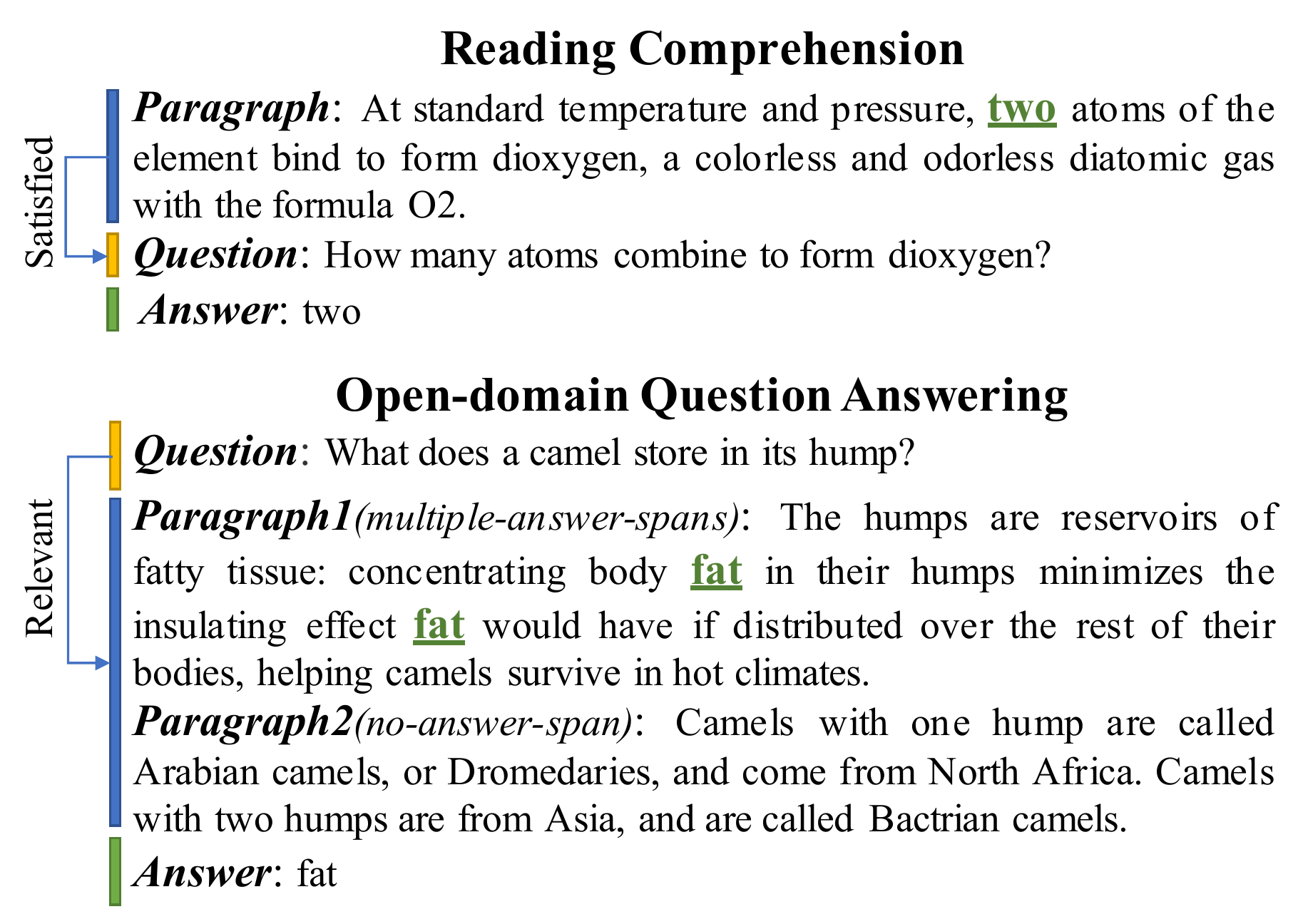}
  \caption{Examples of RC task and OpenQA task.}\label{Fig.example}
\end{figure}

Inspired by the reading comprehension (RC) task flourishing in the area of natural language processing~\cite{wang2016machine,seo2016bidirectional,xiong2016dynamic}, some recent works have viewed OpenQA as an RC task, and directly applied the existing RC models to it~\cite{chen2017reading,joshi2017triviaqa,wang2016machine,clark2017simple}. However, these RC models do not well fit for the OpenQA task.

Firstly, they directly omit the paragraphs without answer string\footnote{The {\em answer string} is a piece of text that can answer the question. If the answer string is obtained in a paragraph as a consecutive text, we call it the {\em answer span}.}. RC task assumes that the given paragraph contains the answer string (Figure~\ref{Fig.example} top), however, it is not valid for the OpenQA task (Figure~\ref{Fig.example} bottom). That\rq{}s because the paragraphs to provide answer for an OpenQA question is collected from a search engine, where each retrieved paragraph is merely relevant to the question. Therefore, it contains many paragraphs without answer string, for instance, in Figure~\ref{Fig.example} Paragraph2. When applying RC models to OpenQA task, we have to omit these paragraphs in the training phase. However, during the inference phase, when model meets one paragraph without answer string, it will pick out a text span as an answer span with high confidence, since RC model has no evidence to justify whether a paragraph contains the answer string.

Secondly, they only consider the first answer span in the paragraph, but omit the remaining rich multiple answer spans. In RC task, the answer and its positions in the paragraph are provided by the annotator in the training data. Therefore RC models only need to consider the unique answer span, e.g.,~in SQuAD~\cite{rajpurkar2016squad}. However, the OpenQA task only provides the answer string as the ground-truth. Therefore, multiple answer spans are detected in the given paragraph, which cannot be considered by the traditional RC models. Take Figure~\ref{Fig.example} as an example, all text spans contain `fat' are treated as answer span, so we detect two answer spans in Paragraph1.

Thirdly, they assume that the start position and end position of an answer span is independent. However, the end position is evidently related with the start position, especially when there are multiple answer spans in a paragraph. Therefore, it may introduce some problems when using such independence assumption. For example, the detected end position may correspond to another answer span, rather than the answer span located by the start position. In Figure~\ref{Fig.example} Paragraph1, `fat in their $\cdots$ insulating effect fat' has a high confidence to be an answer span under independence assumption.

\begin{figure}
  \centering
  \includegraphics[width=0.45\textwidth]{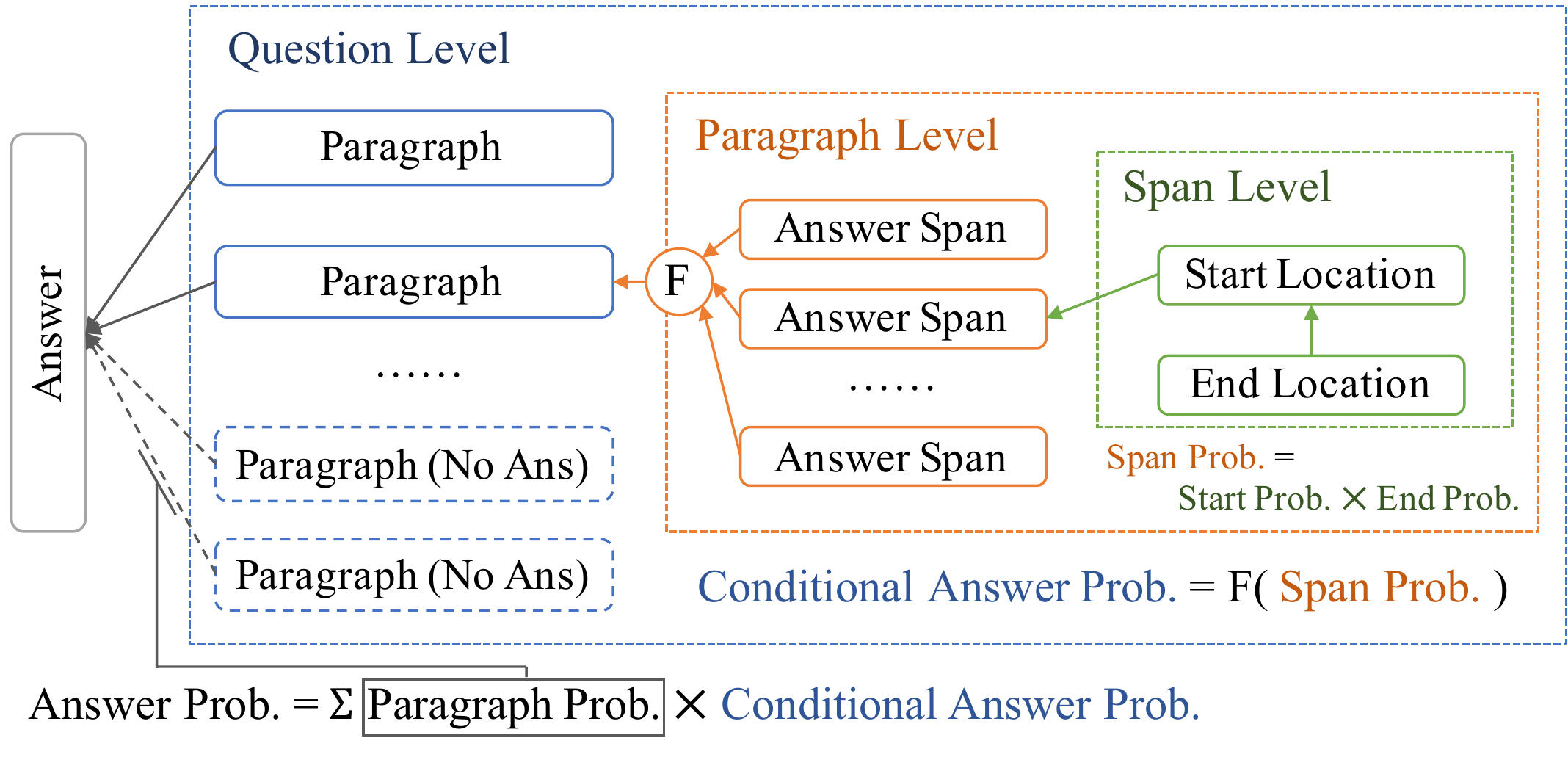}
  \caption{The three hierarchical levels of OpenQA task. }\label{Fig.prob}
\end{figure}

In this paper, we propose a Hierarchical Answer Span Model, named HAS-QA, based on a new three-level probabilistic formulation of OpenQA task, as shown in Figure~\ref{Fig.prob}.

At the question level, the conditional probability of the answer string given a question and a collection of paragraphs, named {\em answer probability}, is defined as the product of the {\em paragraph probability} and {\em conditional answer probability}, based on the law of total probability. 

At the paragraph level, {\em paragraph probability} is defined as the degree to which a paragraph can answer the question. This probability is used to measure the quality of a paragraph and targeted to tackle the first problem mentioned, i.e. identify the useless paragraphs. 
For calculation, we first apply bidirectional GRU and an attention mechanism on the question aware context embedding to obtain a score. Then, we normalize the scores across the multiple paragraphs. In the training phase, we adopt a negative sampling strategy for optimization.
{\em Conditional answer probability} is the conditional probability that a text string is the answer given the paragraph.
Considering multiple answer spans in a paragraph, the {\em conditional answer probability} can be further represented as the aggregation of several {\em span probability}, defined later. In this paper, four types of functions, i.e. HEAD, RAND, MAX and SUM, are used for aggregation.

At the span level, {\em span probability} represents the probability that a text span in a paragraph is the answer span. 
Similarly to previous work~\cite{wang2016machine}, {\em span probability} can be computed as the product of two location probability, i.e.,~{\em location start probability} and {\em location end probability}. Then a conditional pointer network is proposed to model the probabilistic dependences between the start and end positions, by making generation of end position depended on the start position directly, rather than internal representation of start position~\cite{vinyals2015pointer}.

The contributions of this paper include:

1) a probabilistic formulation of the OpenQA task, based on the a three-level hierarchical structure, i.e.~the question level, the paragraph level and the answer span level;
 
2) the proposal of an end-to-end HAS-QA model to implement the three-level probabilistic formulation of OpenQA task (Section~\ref{sec:model}), which tackles the three problems of direct applying existing RC models to OpenQA;
 
3) extensive experiments on QuasarT, TriviaQA and SearchQA datasets, which show that HAS-QA outperforms traditional RC baselines and recent OpenQA baselines.

\section{Related Works}
\label{sec:related-works}
Research in reading comprehension grows rapidly, and many successful RC models have been proposed~\cite{dhingra2017gated,seo2016bidirectional,wang2016machine} in this area. Recently, some works have treated OpenQA task as an RC task and directly applied existing RC models. In this section, we first review the approach of typical RC models, then introduce some recent OpenQA models which are directly based on the RC approach.

RC models typically have two components: context encoder and answer decoder. 
Context encoder is used to obtain the embeddings of questions, paragraphs and their interactions. Most of recent works are based on the attention mechanism and its extensions. The efficient way is to treat the question as a key to attention paragraph \cite{wang2016machine,chen2017reading}. Adding the attention from paragraph to question~\cite{seo2016bidirectional,xiong2016dynamic}, enriches the representations of context encoder. Some works~\cite{wang2017gated,pan2017memen,clark2017simple} find that self-attention is useful for RC task. 
Answer decoder aims to generate answer string based on the context embeddings. There exist two sorts of approaches, generate answer based on the entail word vocabulary~\cite{tan2017s} and retrieve answer from the current paragraph. Almost all works in RC task choose the retrieval-based method. Some of them use two independently position classifiers~\cite{chen2017reading,weissenborn2017fastqa}, the others use the pointer networks~\cite{wang2016machine,seo2016bidirectional,wang2017gated,pan2017memen}. An answer length limitation is applied in these models, i.e. omit the text span longer than 8. We find that relaxing length constrain leads to performance drop.

Some recent works in OpenQA research directly introduce RC model to build a pure data driven pipline.
DrQA~\cite{chen2017reading} is the earliest work that applies RC model in OpenQA task. However, its RC model is trained using typical RC dataset SQuAD~\cite{rajpurkar2016squad}, which turns to be over-confidence about its predicted results even if the candidate paragraphs contain no answer span. 
R${}^3$~\cite{wang2017r} introduces a ranker model to rerank the original paragraph list, so as to improve the input quality of the following RC model. The training data of the RC model is solely limited to the paragraphs containing the answer span and the first appeared answer span location is chosen as the ground truth. 
Shared-Norm~\cite{clark2017simple} applied a shared-norm trick which considers paragraphs without answer span in training RC models. The trained RC model turns to be robust for the useless paragraphs and generates the lower span scores for them. However, it assumes that the start and the end positions of an answer span are independent, which is not suitable for modeling multiple answer spans in one paragraph.

Therefore, we realize that the existing OpenQA models rarely consider the differences between RC and OpenQA task. In this paper, we directly model the OpenQA task based on a probabilistic formulation, in order to identify the useless paragraphs and utilize the multiple answer spans.

\section{Probabilistic Views of OpenQA}
\label{sec:prob-openqa}
In OpenQA task, the question $Q$ and its answer string $A$ are given. Entering question $Q$ into a search engine, top $K$ relevant paragraphs are returned, denote as a list $\mathbf{P} = [P_1,\dots, P_K]$. The target of OpenQA is to find the maximum probability of $P(A|Q, \mathbf{P})$, named {\em answer probability} for short. We can see the following three characteristics of OpenQA:
 
1) we cannot guarantee that paragraph retrieved by search engine contains the answer span for the question, so the paragraphs without answer span have to be deleted when using the above RC models. However, these paragraphs are useful for distinguishing the quality of paragraphs in training. More importantly, the quality of a paragraph plays an important role in determining the {\em answer probability} in the inference phase. It is clear that directly applying RC models fails to meet this requirement.

2) only answer string is provided, while the location of the answer string is unknown. That means there may be many answer spans in the paragraph. It is well known that traditional RC models are only valid for a single answer span. To tackle this problem, the authors of~\cite{joshi2017triviaqa} propose a distantly supervised method to use the \emph{first} exact match location of answer string in the paragraph as the ground-truth answer span. However, this method omit the valuable multiple answer spans information, which may be important for the calculation of the {\em answer probability}. 

3) the start and end positions are coupled together to determine a specific answer span, since there may be multiple answer spans. However, existing RC models usually assume that the start and end positions are independent. 
That\rq{}s because there is only one answer span in the RC scenario. This may introduce serious problem in the OpenQA task. For example, if we do not consider the relations between the start and end position, the end position may be another answer span\rq{}s end position, instead of the one determined by the start position. Therefore, it is not appropriate to assume independence between start and end positions. 
 
In this paper, we propose to tackle the above three problems. Firstly, according to the law of total probability, the {\em answer probability} can be rewritten as the following form.
\begin{equation}
\label{Eq.SQA1} 
  P(A|Q, \mathbf{P})\! =\! \sum_{i=1}^{K} P(P_i|Q, \mathbf{P}) P(A|Q, P_i).
\end{equation}

We name $P(P_i|Q, \mathbf{P})$ and $P(A|Q, P_i)$ as the {\em paragraph probability} and {\em conditional answer probability}, respectively. We can see that the {\em paragraph probability} measures the quality of paragraph $P_i$ across the list $\mathbf{P}$, while the {\em conditional answer probability} measures the probability that string $A$ is an answer string given paragraph $P_i$. 

The {\em conditional answer probability} can be treated as a function of multiple {\em span probabilities} $\{P(L_j(A)|Q, P_i)\}_j$, as shown in Eq~\ref{Eq.SQA2}. 

\begin{equation}
\label{Eq.SQA2}
	\begin{aligned}
		P(A|Q, P_i) &:= \mathcal{F}(\{P(L_j(A)|Q, P_i)\}_j), \\
  			&j \in [1, |\mathcal{L}(A,P_i)|],
	\end{aligned}
\end{equation}
where the aggregation function $\mathcal{F}$ treats a list of spans $\mathcal{L}(A,P_i)$ as input, and $|\mathcal{L}(A,P_i)|$ denotes the number of the text spans contain the string $A$. A proper aggregation function makes use of all the answer spans information in OpenQA task.
Previous work~\cite{joshi2017triviaqa} can be treated as a special case, which uses a function of selecting first match span as the aggregation function $\mathcal{F}$. 

The {\em span probability} $P(L_j(A)|Q, P_i)$ represents the probability that a text span $L_j(A)$ in the paragraph $P_i$ is an answer span. We further decompose it into the product of {\em location start probability} $P(L^s_j(A)|Q, P_i)$ and {\em location end probability} $P(L^e_j(A)|Q, P_i, L^s_j(A))$, shown in Eq~\ref{Eq.SQA3}. 
\begin{equation}
	\label{Eq.SQA3} 
	\begin{aligned}
  	P(L_j(A)|Q, P_i) = &P(L^s_j(A)|Q, P_i) \\
  	\cdot &P(L^e_j(A)|Q, P_i, L^s_j(A)).		
	\end{aligned}
\end{equation}

Some previous work such as DrQA~\cite{chen2017reading} treats them as the two independently position classification tasks, thus $L^{s}(A)$ and $L^{e}(A)$ are modeled by two different functions. Match-LSTM~\cite{wang2016machine} treats them as the pointer networks~\cite{vinyals2015pointer}. The difference is that $L^{e}(A)$ is the function of the hidden state of $L^{s}(A)$, denote as $\mathbf{M^s}$. However, $L^{s}(A)$ and $L^{e}(A)$ are still independent in probabilistic view, because $L^{e}(A)$ depends on the hidden state $\mathbf{M^s}$, not the start position $L^{s}(A)$. In this paper, the span positions $L^s_j(A)$ and $L^e_j(A)$ are determined by the question $Q$ and the paragraph $P_i$. Specially, end position $L^e_j(A)$ is also conditional on start position $L^s_j(A)$ directly. With this conditional probability, we can naturally remove the answer length limitation.

With above formulation, we find that RC task is a special case of OpenQA task, where we set the number of paragraph $K$ to 1, set the {\em paragraph probability} to constant number 1, treat $P(A|Q,P){=}P(L(A)|Q, P)$, $P(L(A)|Q, P){=}P(L^{s}(A)|Q, P)P(L^{e}(A)|Q, P)$, where $P$ is the idealized paragraph that contain the answer string $A$, and the right position $L(A)$ is also known. 

\section{HAS-QA Model} 
\label{sec:model}
In this section, we propose a Hierarchical Answer Span Model (HAS-QA) for OpenQA task, based on the probabilistic view of OpenQA in Section~\ref{sec:prob-openqa}. HAS-QA has four components: question aware context encoder, conditional span predictor, multiple spans aggregator and paragraph quality estimator. We will introduce them one by one.

\subsection{Question Aware Context Encoder}
\label{sec:contexts-net}
The question aware context embeddings $\mathbf{C}$ is generated by the context encoder, while HAS-QA do not limit the use of context encoder.
We choose a simple but efficient context encoder in this paper. It takes advantage of previous works~\cite{clark2017simple,wang2016machine}, which contains the character-level embedding enhancement, the bi-directional attention mechanism~\cite{seo2016bidirectional} and the self-attention mechanism~\cite{wang2017gated}. We briefly describe the process below~\footnote{For more detailed computational steps, see reference paper \cite{clark2017simple}.}. 

\textbf{Word Embeddings}: use size 300 pre-trained GloVe~\cite{pennington2014glove} word embeddings.

\textbf{Char Embeddings}: encode characters in size 20, which are learnable. Then obtain the embedding of each word by convolutional layer and max pooling layer. 

\textbf{Context Embeddings}: concatenate word embeddings and char embeddings, and apply bi-directional GRU~\cite{cho2014learning} to obtain the context embeddings. Both question and paragraph get their own context embeddings.

\textbf{Question Aware Context Embeddings}: use bi-directional attention mechanism from the BiDAF~\cite{seo2016bidirectional} to build question aware context embeddings. Additionally, we subsequently apply a layer of self-attention to get the final question aware context embeddings.

After the processes above, we get the final question aware context embeddings, denoted $\mathbf{C} \in \mathbb{R}^{n \times r}$, where $n$ is the length of the paragraph and $r$ is size of the embedding.

\subsection{Conditional Span Predictor} 
\label{sec:cond-pointer-net}
Conditional span predictor defines the {\em span probability} for each text span in a paragraph using a conditional pointer network.

We first review the answer decoder in traditional RC models. It mainly has two types: two independently position classifiers (IndCls) and the pointer networks (PtrNet). Both of these approaches generate a distribution of start position $\mathbf{p^s} \in \mathbb{R}^n$ and a distribution of end position $\mathbf{p^e} \in \mathbb{R}^n$, where $n$ is the length of the paragraph. 
Starting from the context embeddings $\mathbf{C}$, two intermedia representations $\mathbf{M^s} \in \mathbb{R}^{n \times 2d}$ and $\mathbf{M^e} \in \mathbb{R}^{n \times 2d}$ are generated using two  bidirectional GRUs with the output dimension $d$.
\begin{align}
	\label{Eq.pointer} \mathbf{M^s} &= \mathrm{BiGRU}(\mathbf{C})\\
	\label{Eq.pointer_indcls} \textrm{IndCls:}\; \mathbf{M^e} &= \mathrm{BiGRU}(\mathbf{C}), \\
	\label{Eq.pointer_ptrnet} \textrm{PtrNet:}\; \mathbf{M^e} &= \mathrm{BiGRU}([\mathbf{C}, \mathbf{M^s}]).
\end{align}

Then an additional Softmax function is used to generate the final positional distributions,
\begin{equation} \label{Eq.ps}
	\begin{aligned}
		&\mathbf{p^s}\! =\! \mathrm{softmax}(\mathbf{M^s}w_s), \\
		&\mathbf{p^e}\! =\! \mathrm{softmax}(\mathbf{M^e}w_e).
	\end{aligned}
\end{equation}
where $w_s, w_e \in \mathbb{R}^{2d}$ denotes the linear transformation parameters.

As mentioned in Section~\ref{sec:prob-openqa}, IndCls and PtrNet both treat start and end position as probabilistic independent. Given the independent start and end positions can not distinguish the different answer spans in a paragraph properly, so it is necessary to build a conditional model for them.
Therefore, we proposed a conditional pointer network which directly feed the start position to the process of generating the end position:
\begin{equation} \label{Eq.end_prob}
	\begin{aligned}
		\mathbf{M^e_j} &= \mathrm{BiGRU}([\mathbf{C}, \mathbf{M^s}, \mathrm{OneHot}(L^s_j)]), \\
		\mathbf{p^e_j} &= \mathrm{softmax}(\mathbf{M^e_j}w_e),
	\end{aligned}
\end{equation}
where $L^s_j$ denotes the start position selected from the start positional distribution $\mathbf{p^s}$ and $\mathrm{OneHot}(\cdot)$ denotes the transformation from a position index to an one-hot vector. 

In the training phase, we are given the start and end positions of each answer span, denote as $L^s_j$ and $L^e_j$. The {\em span probability} is:
\begin{equation} \label{Eq.span_prob}
	P(L_j(A)|Q, P_i) = s_j = \mathbf{p^s}[L^s_j] \cdot \mathbf{p^e_j}[L^e_j].
\end{equation}

In the inference phase, we first select the start position $L^s_j$ from the start distribution $\mathbf{p^s}$. Then we yield its corresponding end distribution $\mathbf{p^e_j}$ using Eq~\ref{Eq.end_prob}, and select the end position $L^e_j$ from it. Finally, we get the {\em span probability} using Eq~\ref{Eq.span_prob}.

\subsection{Multiple Spans Aggregator}
\label{sec:agg-net}
Multiple span aggregator is used to build the relations among multiple answer spans and outputs the {\em conditional answer probability}. In this paper, we design four types of aggregation functions $\mathcal{F}$:
\begin{equation} \label{Eq.agg_func}
	\begin{aligned}
		&\textrm{HEAD:} \; P(A|Q, P_i) = s_1 \\
		&\textrm{RAND:} \; P(A|Q, P_i) = \textrm{Random}(s_j) \\
		&\textrm{MAX:} \;\;\; P(A|Q, P_i) = \max_j\nolimits (s_j) \\
		&\textrm{SUM:} \;\;\; P(A|Q, P_i) = \sum_j\nolimits (s_j) \\
	\end{aligned}
\end{equation}
where $s_j$ denotes the {\em span probability} defined in Eq~\ref{Eq.span_prob}, $s_1$ denotes the first match answer span and $\textrm{Random}$ denotes a stochastic function for randomly choosing an answer span. 

Different aggregation functions represent different assumptions about the distribution of the oracle answer spans in a paragraph. The oracle answer span represents the answer of the question that can be merely determined by its context, e.g. in Figure~\ref{Fig.example}, the first answer span `fat' is the oracle answer span, while the second one is not, because we could retrieval the answer directly, if we have read `concentrating body fat in their humps'.  

\textbf{HEAD} operation simply chooses the first match {\em span probability} as the {\em conditional answer probability}, which simulates the answer preprocessing in previous works~\cite{wang2017r,joshi2017triviaqa}. This function only encourages the first match answer span as the oracle, while punishes the others. It can be merely worked in a paragraph with definition, such as first paragraph in WikiPedia.

\textbf{RAND} operation randomly chooses a {\em span probability} as the {\em conditional answer probability}. This function assumes that all answer spans are equally important, and must be treated as oracle. However, balancing the probabilities of answer spans is hard. It can be used in paraphrasing answer spans appear in a list.

\textbf{MAX} operation chooses the maximum {\em span probability} as the {\em conditional answer probability}. This function assumes that only one answer span is the oracle. It can be used in a noisy paragraph, especially for those retrieved by a search engine.

\textbf{SUM} operation sums all the {\em span probabilities} as the {\em conditional answer probability}. This function assumes that one or more answer spans are the oracle. It can be used in a broad range of scenarios, for its relatively weak assumption.

In the training phase, all annotated answer spans contain the same answer string $A$, we directly apply the Eq~\ref{Eq.agg_func} to obtain the {\em conditional answer probability} in paragraph level.

In the inference phase, we treat the top $K$ {\em span probabilities} $s_j$ as the input of the aggregation function. However, we have to check all possible start and end positions to get the precise top $K$ {\em span probabilities}. Instead, we use a beam search strategy~\cite{sutskever2014sequence} which only consider the top $K_1$ start positions and the top $K_2$ end positions, where $K_1 K_2 \ge K$. 
Different {\em span probabilities} $s_j$ represent variance answer strings $A_t$. Following the definition in Eq~\ref{Eq.agg_func}, we group them by different answer strings respectively.

\subsection{Paragraph Quality Estimator}
\label{sec:neg-sample}
Paragraph quality estimator takes the useless paragraphs into consideration, which implements the {\em paragraph probability} $P(P_i|Q, \mathbf{P})$ directly. 

Firstly, we use an attention-based network to generate a quality score, denotes as $\hat{q}_i$, in order to measure the quality of the given paragraph $P_i$.

\begin{equation} \label{Eq.quality_score}
	\begin{aligned}
		&\mathbf{M^c} = \textrm{BiGRU}(\mathbf{C}),\\
		&\hat{q}_i = (\mathbf{M^c}^{\top} \cdot \mathbf{p^s}) \cdot w_c.
	\end{aligned}
\end{equation}
where $\mathbf{M^c} \in \mathbb{R}^{n \times 2d}$ is the intermedia representation obtained by applying bidirectional GRU on the context embedding $\mathbf{C}$. Then, let start distribution $\mathbf{p^s} \in \mathbb{R}^n$ as a key to attention $\mathbf{M^c}$ and transform it to 1-d value using weight $w_c \in \mathbb{R}^{2d}$. Finally, we get the quality score $\hat{q}_i$. {\em Paragraph probabilities} $P(P_i|Q, \mathbf{P})$ are generated by normalizing across $\mathbf{P}$,
\begin{equation} \label{Eq.quality_prob}
	P(P_i|Q, \mathbf{P})\! =\! q_i =\! \frac{\exp(\hat{q}_i)}{\sum_{P_j \in \mathbf{P}} \exp(\hat{q}_j)}.
\end{equation}

In the training phase, we conduct a negative sampling strategy with one negative sample, for efficient training. Thus a pair of paragraphs, $P^+$ as positive and  $P^-$ as negative, are used to approximate $q^+ \approx P(P^+|Q, [P^+, P^-])$ and $q^- \approx P(P^-|Q, [P^+, P^-])$.

In the inference phase, the probability $q_i$ is obtained by normalizing across all the retrieved paragraphs $\mathbf{P}$.

\begin{algorithm}[h]
\caption{HAS-QA Model in Training Phase} 
\label{alg:training} 
\begin{algorithmic}[1] 
\Require
$Q$: question; 
$A$: answer string;

$\mathbf{P}$: retrieved paragraphs; 

\Ensure 
$\mathcal{L}$: loss function

\For {$P^+$, $P^-$ in $\mathbf{P}$}:
	\State Get answer locations $\mathbf{L^s}$, $\mathbf{L^e}$ for $P^+$;
	\State Get the context embedding $\mathbf{C}$;
	\State Compute $\mathbf{p^s}$; \hfill (Eq~\ref{Eq.ps})
	\For {$L^s_j, L^e_j$ in $\mathbf{L^s}, \mathbf{L^e}$}:
		\State $p^s_j \gets \mathbf{p^s}[L^s_j]$;
		\State Compute $\mathbf{p^e_j}$; \hfill (Eq~\ref{Eq.end_prob})
		\State $p^e_j \gets \mathbf{p^e_j}[L^e_j]$; 
		\State $s_j \gets p^s_j p^e_j$;
	\EndFor
	\State Apply function: $p^+ \gets \mathcal{F}(\{s_j\})$;
	\State Compute $q^+$ in $[P^+, P^-]$; \hfill (Eq~\ref{Eq.quality_score}, Eq~\ref{Eq.quality_prob})
	\State $\mathcal{L}_i \gets -(\log(q^+) +\log(p^+))$;
\EndFor 
\State $\mathcal{L} \gets \mathrm{Avg}(\{\mathcal{L}_i\})$.

\end{algorithmic} 
\end{algorithm}

\begin{algorithm}[h]
\caption{HAS-QA Model in Inference Phase} 
\label{alg:inference} 
\begin{algorithmic}[1] 
\Require
$Q$: question; 
$\mathbf{P}$: retrieved paragraphs; 
\Ensure 
$A_{best}$: answer string

\For {$P_i$ in $\mathbf{P}$}:
	\State Get the context embedding $\mathbf{C}$;
	\State Compute $\mathbf{p^s}$; \hfill (Eq~\ref{Eq.ps})
	\For {$L^s_j$ in Top-$K_1$ $\mathbf{p^s}$}:
		\State $p^s_j \gets \mathbf{p^s}[L^s_j]$;
		\State Compute $\mathbf{p^e_j}$; \hfill (Eq~\ref{Eq.end_prob})
		\For {$L^e_{jk}$ in Top-$K_2$ $\mathbf{p^e_j}$}:
			\State $p^e_{jk} \gets \mathbf{p^e_j}[L^e_{jk}]$; 
			\State $s_{jk} \gets p^s_j p^e_{jk}$;
		\EndFor
	\EndFor
	\State Group $s_{jk}$ by extracted answer string $A_t$;
	\State Apply function: $p^{A_t}_i \gets \mathcal{F}(\{s_{jk}\}_{A_t})$;
	\State Compute $\hat{q}_i$; \hfill (Eq~\ref{Eq.quality_score})
\EndFor 
 
\State Normalize $\{\hat{q}_i\}$ get $\{q_i\}$; \hfill (Eq~\ref{Eq.quality_prob})
\State $S(A_t) \gets \sum_i q_i \cdot p^{A_t}_i$;
\State $A_{best} \gets \arg\max(S(A_t))$.

\end{algorithmic} 
\end{algorithm}

Above all, we describe our model with Algorithm~\ref{alg:training} in the training phase and Algorithm~\ref{alg:inference} in the inference phase.

\section{Experiments}
\label{sec:experiments}

\begin{table}
    \centering
	\small
	\begin{tabular}{l r r}
		\hline
		Dataset & Neg Para. Ratio & Avg Ans. Span Count\\
		\hline
		\hline
		QuasarT  & 1.21\% & 5.09 \\
		TriviaQA & 37.24\% & 4.20 \\ 
		SearchQA & 25.06\% & 6.80 \\ 
		\hline
	\end{tabular}
	\caption{The negative paragraph ratio and average answer span count are statistic on three datasets, in order to illustrate the problems mentioned above in OpenQA task.}
	\label{Table.data_statistic}
\end{table}

\begin{table*}
    \centering
	\begin{tabular}{l r r r r r r}
		\hline
		&\multicolumn{2}{c}{QuasarT} & \multicolumn{2}{c}{TriviaQA} & \multicolumn{2}{c}{SearchQA} \\
		Model & EM & F1 & EM & F1 & EM & F1\\
		\hline
		\hline
		GA~\cite{dhingra2017gated}  		
					& 0.264 & 0.264 
				   	& - & - 
				   	& - & -\\ 
		BiDAF~\cite{seo2016bidirectional} 	 	
					& 0.259 & 0.285  
					& 0.411 & 0.474
					& 0.286 & 0.346\\ 
		AQA~\cite{buck2017ask} 		
					& - & - 
					& - & -
					& 0.387 & 0.456\\
		\hline 
		DrQA~\cite{chen2017reading}			
					& 0.377 & 0.445
					& 0.323 & 0.383
					& 0.419 & 0.487\\
		R${}^3$~\cite{wang2017r}	
					& 0.353 & 0.417  % & 0.065 & 0.149
					& 0.473 & 0.537 
					& 0.490 & 0.553\\ 
		Shared-Norm~\cite{clark2017simple}
					& 0.386 & 0.454 % quasar-0122-094033
					& 0.613 & 0.672 % debug-1231-053013
					& 0.598 & 0.671 \\ %searchqa-0205-065301 
		\hline
		HAS-QA (MAX Ans. Span) 
					& \textbf{0.432} & \textbf{0.489} % quasar-0122-100527
					& \textbf{0.636} & \textbf{0.689} % debug-0119-052919_400_16_3
					& \textbf{0.627} & \textbf{0.687} \\ %searchqa-0206-035521 
		\hline
	\end{tabular}
	\caption{Experimental results on OpenQA datasets QuasarT, TriviaQA and SearchQA. EM: Exact Match.}
	\label{Table.main}
\end{table*}

\subsection{Datasets}
We evaluate our model on three OpenQA datasets, QuasarT~\cite{dhingra2017quasar}, TriviaQA~\cite{joshi2017triviaqa} and SearchQA~\cite{dunn2017searchqa}. 

\textbf{QuasarT}\footnote{https://github.com/bdhingra/quasar}: 
consists of 43k open-domain trivia questions whose answers obtained from various internet sources. ClueWeb09~\cite{callan2009clueweb09} serves as the background corpus for providing evidences paragraphs. We choose the Long version, which is truncated to 2048 characters and 20 paragraphs for each question.

\textbf{TriviaQA}\footnote{http://nlp.cs.washington.edu/triviaqa/}: 
consists of 95k open-domain question-answer pairs authored by trivia enthusiasts and independently gathered evidence documents from Bing Web Search and Wikipedia, six per question on average. We focus on the open domain setting contains unfiltered documents.

\textbf{SearchQA}\footnote{https://github.com/nyu-dl/SearchQA}:
is based on a Jeopardy! questions and collects about top 50 web page snippets from Google search engine for each question.

As we can see in Table~\ref{Table.data_statistic}, there exist amounts of negative paragraphs which contains no answer span, especially in TriviaQA and SearchQA. For all datasets, more than 4 answer spans averagely obtained per paragraph. These statistics illustrate that problems mentioned above exist in OpenQA datasets.

\subsection{Experimental Settings}
For RC baseline models GA \cite{dhingra2017gated}, BiDAF \cite{seo2016bidirectional} and AQA \cite{buck2017ask}, their experimental results are collected from published papers \cite{dunn2017searchqa,joshi2017triviaqa}. 

The DrQA \cite{chen2017reading}, R${}^3$ \cite{wang2017r} and Shared-Norm \cite{clark2017simple} are evaluated using their released code\footnote{DrQA: https://github.com/facebookresearch/DrQA. \\R${}^3$: https://github.com/shuohangwang/mprc. \\Shared-Norm: https://github.com/allenai/document-qa.}.

Our model~\footnote{The code will be released at https://gitlab.com/pl8787/has-qa.} adopts the same data preprocessing and question context encoder presented in \cite{clark2017simple}. 
In training step, we use the Adadelta optimizer~\cite{zeiler2012adadelta} with the batch size of 30, and we choose the model performed the best on develop set~\footnote{QuasarT and SearchQA have official develop set and test set, while TriviaQA's test set is unknown, thus we split a develop set from train set and evaluate on official develop set. }. The hidden dimension of GRU is 200, and the dropout ratio is 0.8. We use 300 dimensional word embeddings pre-trained by GloVe (released by \cite{pennington2014glove}) and do not fine-tune in training step. Additionally, 20 dimensional character embeddings are left as learnable parameters.
In inference step, for baseline models we set the answer length limitation to 8, while for our models it is unlimited. We analyze different answer length limitation settings in the Section~\ref{sec:ana_conptr}. The parameters of beam search are $K_1=3$ and $K_2=1$. 

\subsection{Overall Results}
The experimental results on three OpenQA datasets are shown in Table~\ref{Table.main}. 
It concludes as follow:

1) HAS-QA outperforms traditional RC baselines with a large gap, such as GA, BiDAF, AQA listed in the first part. For example, in QuasarT, it improves 16.8\% in EM score and 20.4\% in F1 score.
As RC task is just a special case of OpenQA task. Some experiments on standard SQuAD dataset(dev-set)~\cite{rajpurkar2016squad} show that HAS-QA yields EM/F1:0.719/0.798, which is comparable with the best released single model Reinforced Mnemonic Reader~\cite{hu2017reinforced} in the leaderboard (dev-set) EM/F1:0.721/0.816. Our performance is slightly worse because Reinforced Mnemonic Reader directly use the accurate answer span, while we use multiple distantly supervised answer spans. That may introduce noises in the setting of SQuAD, since only one span is accurate. 

2) HAS-QA outperforms recent OpenQA baselines, such as DrQA, R${}^3$ and Shared-Norm listed in the second part. For example, in QuasarT, it improves 4.6\% in EM score and 3.5\% in F1 score.  

\subsection{Model Analysis}
In this subsection, we analyze our model by answering the following fine-grained analytic questions: 

1) What advantages does HAS-QA have via modeling answer span using the conditional pointer network? 

2) How much does HAS-QA gain from modeling multiple answer spans in a paragraph? 

3) How does the paragraph quality work in HAS-QA?

The following three parts are used to answer these questions respectively.

\subsubsection{Effects of Conditional Pointer Networks}\label{sec:ana_conptr}
In order to demonstrate the effect of the conditional pointer networks, we compare Shared-Norm, which uses pointer networks, with our model. Then, we gradually remove the answer length limitation, from restricting 4 words to 128 words until no limitation (denote as $\infty$). Finally, we draw the tendency of the EM performance and average predicted answer length according to the different answer length limitations.

\begin{figure} 

	\begin{minipage}[t]{0.5\linewidth}
		\centerline{\includegraphics[width=1\linewidth]{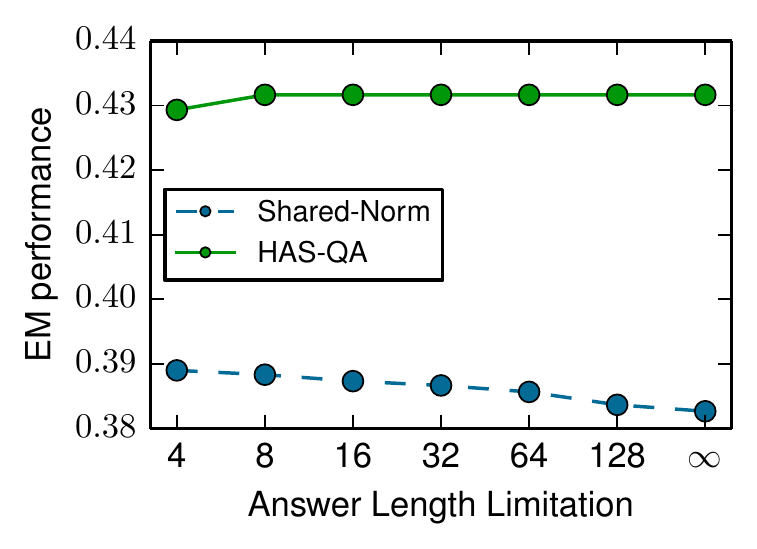}}
	\end{minipage}%
	\begin{minipage}[t]{0.5\linewidth}
		\centerline{\includegraphics[width=1\linewidth]{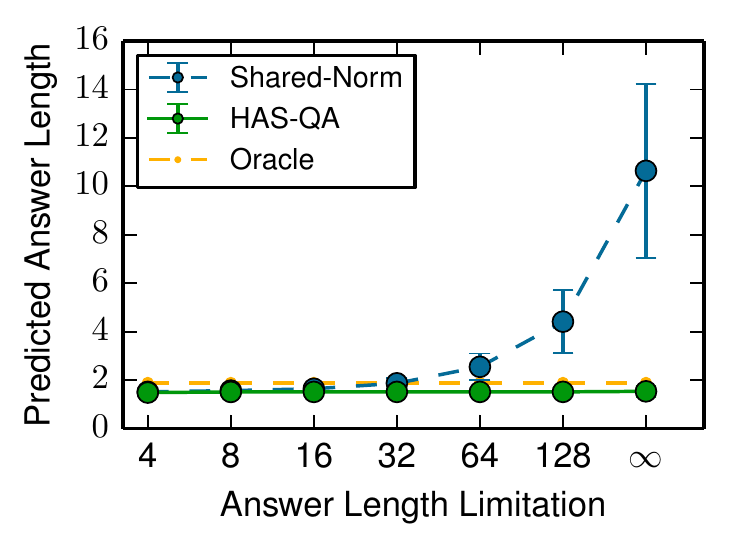}}
	\end{minipage}
	\begin{minipage}[t]{1\linewidth}
		\centerline{\includegraphics[width=1\linewidth]{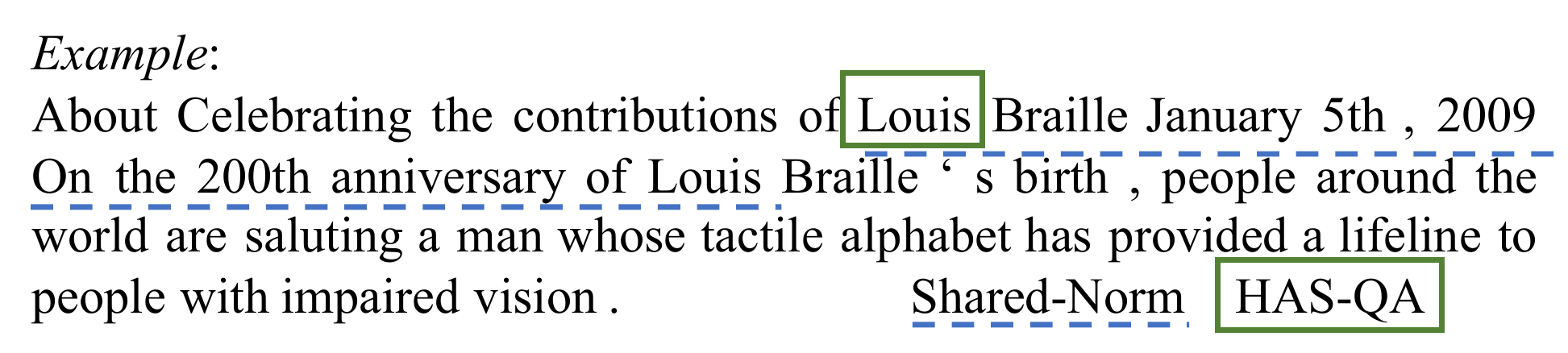}}
	\end{minipage}
	
	\caption{Results of Shared-Norm and HAS-QA on QuasarT. TopLeft: EM performance against answer length limitation, TopRight: predicted answer length against answer length limitation, Bottom: an example of a paragraph and the predicted answer spans of two models.}\label{Fig.AnsLen}
\end{figure}

As shown in Figure~\ref{Fig.AnsLen} (TopLeft), the performance of Shared-Norm decreases when removing the answer length limitation, while the performance of HAS-QA first increases then becomes stable. 
In Figure~\ref{Fig.AnsLen} (TopRight), we find that the average predicted answer length increases in Shared-Norm when removing the answer length limitation. However, our model stably keeps average about 1.8 words, where the oracle average answer length is about 1.9 words. 
Example in Figure~\ref{Fig.AnsLen} (Bottom) illustrates that start/end pointers in Shared-Norm search their own optimal positions independently, such as two `Louis' in paragraph. It leads to an unreasonable answer span prediction.

\subsubsection{Effects of Multiple Spans Aggregation}
The effects of utilizing multiple answer spans lay into two aspects, 1) choose the aggregation functions in training phase, and 2) select the parameters of beam search in inference phase.

In the training phase, we evaluate four types of aggregation functions introduced in Section~\ref{sec:agg-net}.
The experimental results on QuasarT dataset, shown in Table~\ref{Table.agg_func}, demonstrate the superiority of SUM and MAX operations. They take advantages of using multiple answer spans for training and improve about 6\% - 10\% in EM comparing to the HEAD operation. The performance of MAX operation is a little better than the SUM operation.
The failure of RAND operation, mainly comes down to the conflicting training samples. Therefore, simple way to make use of multiple answer spans may not improve the performance. 

\begin{table}
    \centering
	\small
	\begin{tabular}{l r r}
		\hline
		Model & EM & F1\\
		\hline
		\hline
		HAS-QA (HEAD Ans. Span) & 0.372 & 0.425 \\ % quasar-0125-064654 
		HAS-QA (RAND Ans. Span) & 0.341 & 0.394 \\ % quasar-0123-224533
		HAS-QA (SUM Ans. Span) & 0.423 & 0.484 \\ % quasar-0123-112117
		HAS-QA (MAX Ans. Span) & 0.432 & 0.489 \\ % quasar-0127-084213 
		\hline
	\end{tabular}
	\caption{Results on QuasarT with different types of aggregation functions ($K_1=3, K_2=1$).}
	\label{Table.agg_func}
\end{table}

In the inference phase, Table~\ref{Table.beam_size} shows the effects of parameters in beam search. We find that the larger $K_1$ yields the better performance, while $K_2$ seems irrelevant to the performance. As a conclusion, we choose the parameters $K_1=3, K_2=1$ to balance the performance and the speed.

\subsubsection{Effects of Paragraph Quality}
The {\em paragraph probability} is efficient to measure the quality of  paragraphs, especially for that containing useless paragraphs. 

\begin{figure} 
	\begin{minipage}[t]{0.65\linewidth}
	\centering
	\includegraphics[width=1\linewidth]{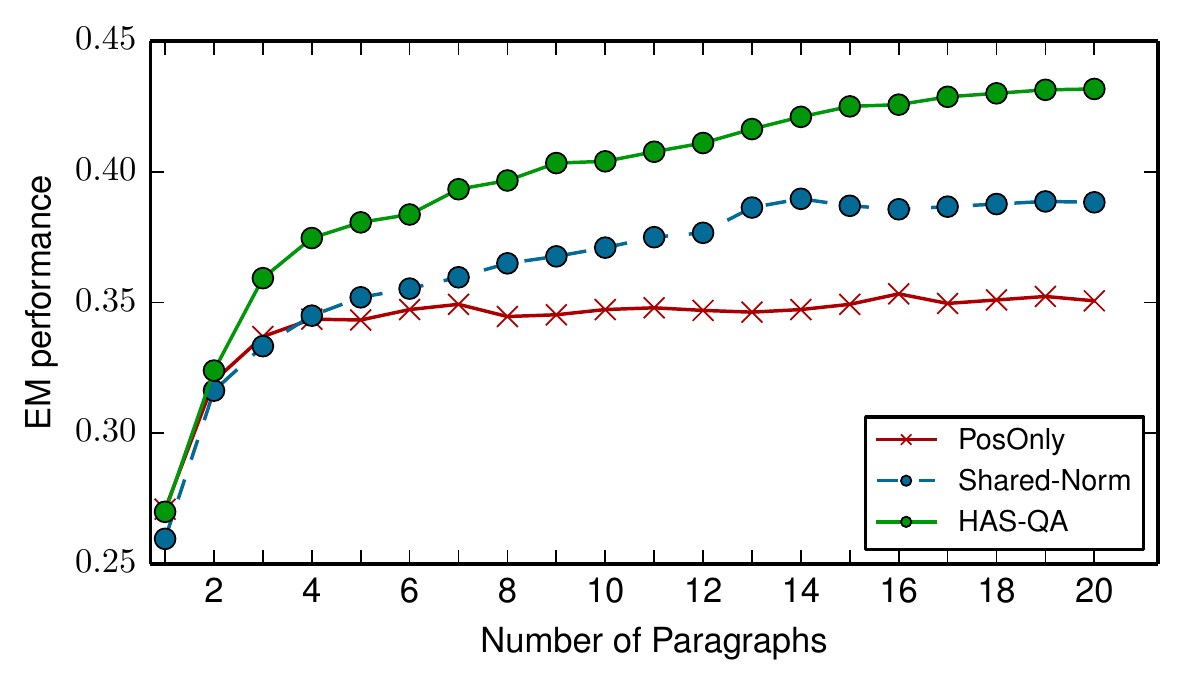}
	%\caption{fig1}
	\label{fig:side:a}
	\end{minipage}%
	\begin{minipage}[t]{0.35\linewidth}
	\centering
	\includegraphics[width=1\linewidth]{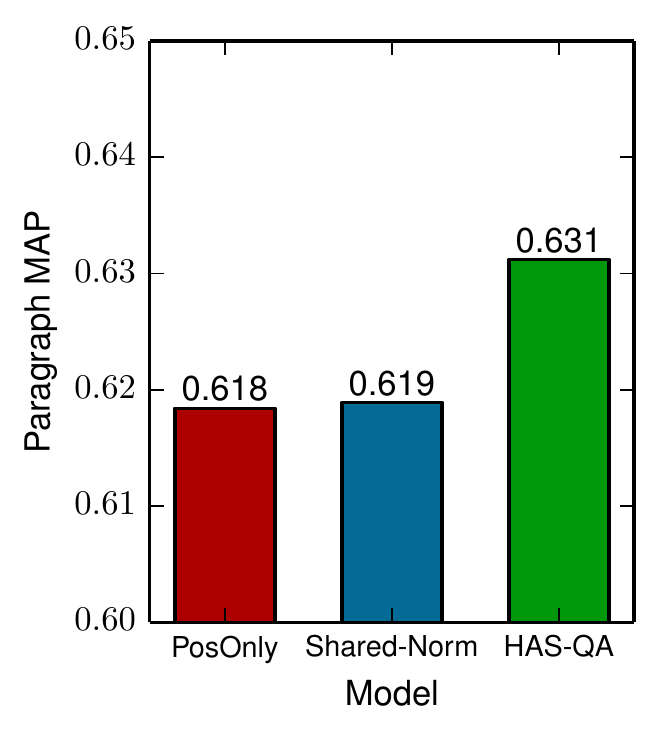}
	%\caption{fig2}
	\label{fig:side:b}
	\end{minipage}
	
	\caption{Results of PosOnly, Shared-Norm and HAS-QA on QuasarT. Left: EM performance against number of paragraphs, Right: paragraph MAP on different models.}\label{Fig.ParaQuality}
\end{figure}

Figure~\ref{Fig.ParaQuality} (Left) shows that with the increasing number of given paragraphs which ordered by the rank of a search engine, EM performance of HAS-QA sustainably grows. However, EM performance of Shared-Norm stops increasing at about 15 paragraphs and our model without paragraph quality (denotes PosOnly) stops increasing at about 5 paragraphs. So that with the help of {\em paragraph probability}, model performance can be improved by adding more evidence paragraphs.

\begin{table}
    \centering
	\small
	\begin{tabular}{c r r c r r}
		\hline
		$K_1$-$K_2$ & EM & F1 &
		$K_1$-$K_2$ & EM & F1\\
		\hline
		\hline
		1-1 & 0.428 & 0.483 &
		1-1 & 0.428 & 0.483 \\
		1-3 & 0.428 & 0.484 &
		3-1 & 0.432 & 0.489 \\
		1-5 & 0.428 & 0.484 &
		5-1 & 0.431 & 0.488 \\
		3-3 & 0.431 & 0.489 &
		5-5 & 0.431 & 0.489 \\
		\hline
	\end{tabular}
	\caption{Results on QuasarT with different beam search parameters $K_1$-$K_2$.}
	\label{Table.beam_size}
\end{table}

We also evaluate the Mean Average Precision (MAP) score between the predicted scores and the label whether a paragraph contains answer spans (Figure~\ref{Fig.ParaQuality} (Right)). The {\em paragraph probability} in our model outperforms PosOnly and Shared-Norm, so that it can rank the high quality paragraphs in the front of the given paragraph list.

\section{Conclusions}\label{sec:conclusion}
In this paper, we point out three distinct characteristics of OpenQA, which make it inappropriate to directly apply existing RC models to this task. 
In order to tackle these problems, we first propose a new probabilistic formulation of OpenQA, where the {\em answer probability} is written as the question, paragraph and span, three-level structure. In this formulation, RC can be treated as a special case. 
Then, Hierarchical Answer Spans Model (HAS-QA) is designed to implement this structure. Specifically, a paragraph quality estimator makes it robust for the paragraphs without answer spans; a multiple span aggregator points out that it is necessary to combine the contributions of multiple answer spans in a paragraph, and a conditional span predictor is proposed to model the dependence between the start and end positions of each answer span. 
Experiments on public OpenQA datasets, including QuasarT, TriviaQA and SearchQA, show that HAS-QA significantly outperforms traditional RC baselines and recent OpenQA baselines.

%ACKNOWLEDGMENTS are optional
\section*{Acknowledgments}
This work was funded by the National Natural Science Foundation of China (NSFC) under Grants No. 61773362, 61425016, 61472401, 61722211, and 61872338, the Youth Innovation Promotion Association CAS under Grants No. 20144310, and 2016102, and the National Key R\&D Program of China under Grants No. 2016QY02D0405. 

\bibliography{has-qa.bib}

\begin{thebibliography}{}

\bibitem[\protect\citeauthoryear{Berant \bgroup et al\mbox.\egroup
  }{2013}]{berant2013semantic}
Berant, J.; Chou, A.; Frostig, R.; and Liang, P.
\newblock 2013.
\newblock Semantic parsing on freebase from question-answer pairs.
\newblock In {\em Proceedings of the 2013 Conference on Empirical Methods in
  Natural Language Processing},  1533--1544.

\bibitem[\protect\citeauthoryear{Buck \bgroup et al\mbox.\egroup
  }{2017}]{buck2017ask}
Buck, C.; Bulian, J.; Ciaramita, M.; Gesmundo, A.; Houlsby, N.; Gajewski, W.;
  and Wang, W.
\newblock 2017.
\newblock Ask the right questions: Active question reformulation with
  reinforcement learning.
\newblock {\em arXiv preprint arXiv:1705.07830}.

\bibitem[\protect\citeauthoryear{Callan \bgroup et al\mbox.\egroup
  }{2009}]{callan2009clueweb09}
Callan, J.; Hoy, M.; Yoo, C.; and Zhao, L.
\newblock 2009.
\newblock Clueweb09 data set.

\bibitem[\protect\citeauthoryear{Chen \bgroup et al\mbox.\egroup
  }{2017}]{chen2017reading}
Chen, D.; Fisch, A.; Weston, J.; and Bordes, A.
\newblock 2017.
\newblock Reading wikipedia to answer open-domain questions.
\newblock In {\em Proceedings of the 55th Annual Meeting of the Association for
  Computational Linguistics (Volume 1: Long Papers)}, volume~1,  1870--1879.

\bibitem[\protect\citeauthoryear{Cho \bgroup et al\mbox.\egroup
  }{2014}]{cho2014learning}
Cho, K.; van Merrienboer, B.; Gulcehre, C.; Bahdanau, D.; Bougares, F.;
  Schwenk, H.; and Bengio, Y.
\newblock 2014.
\newblock Learning phrase representations using rnn encoder--decoder for
  statistical machine translation.
\newblock In {\em Proceedings of the 2014 Conference on Empirical Methods in
  Natural Language Processing (EMNLP)},  1724--1734.

\bibitem[\protect\citeauthoryear{Clark and Gardner}{2018}]{clark2017simple}
Clark, C., and Gardner, M.
\newblock 2018.
\newblock Simple and effective multi-paragraph reading comprehension.
\newblock In {\em Proceedings of the 56th Annual Meeting of the Association for
  Computational Linguistics (Volume 1: Long Papers)}, volume~1,  845--855.

\bibitem[\protect\citeauthoryear{Dhingra \bgroup et al\mbox.\egroup
  }{2017}]{dhingra2017gated}
Dhingra, B.; Liu, H.; Yang, Z.; Cohen, W.; and Salakhutdinov, R.
\newblock 2017.
\newblock Gated-attention readers for text comprehension.
\newblock In {\em Proceedings of the 55th Annual Meeting of the Association for
  Computational Linguistics (Volume 1: Long Papers)}, volume~1,  1832--1846.

\bibitem[\protect\citeauthoryear{Dhingra, Mazaitis, and
  Cohen}{2017}]{dhingra2017quasar}
Dhingra, B.; Mazaitis, K.; and Cohen, W.~W.
\newblock 2017.
\newblock Quasar: Datasets for question answering by search and reading.
\newblock {\em arXiv preprint arXiv:1707.03904}.

\bibitem[\protect\citeauthoryear{Dunn \bgroup et al\mbox.\egroup
  }{2017}]{dunn2017searchqa}
Dunn, M.; Sagun, L.; Higgins, M.; Guney, U.; Cirik, V.; and Cho, K.
\newblock 2017.
\newblock Searchqa: A new q\&a dataset augmented with context from a search
  engine.
\newblock {\em arXiv preprint arXiv:1704.05179}.

\bibitem[\protect\citeauthoryear{Ferrucci \bgroup et al\mbox.\egroup
  }{2010}]{ferrucci2010building}
Ferrucci, D.; Brown, E.; Chu-Carroll, J.; Fan, J.; Gondek, D.; Kalyanpur,
  A.~A.; Lally, A.; Murdock, J.~W.; Nyberg, E.; Prager, J.; et~al.
\newblock 2010.
\newblock Building watson: An overview of the deepqa project.
\newblock {\em AI magazine} 31(3):59--79.

\bibitem[\protect\citeauthoryear{Hu \bgroup et al\mbox.\egroup
  }{2017}]{hu2017reinforced}
Hu, M.; Peng, Y.; Huang, Z.; Qiu, X.; Wei, F.; and Zhou, M.
\newblock 2017.
\newblock Reinforced mnemonic reader for machine reading comprehension.
\newblock {\em arXiv preprint arXiv:1705.02798}.

\bibitem[\protect\citeauthoryear{Joshi \bgroup et al\mbox.\egroup
  }{2017}]{joshi2017triviaqa}
Joshi, M.; Choi, E.; Weld, D.; and Zettlemoyer, L.
\newblock 2017.
\newblock Triviaqa: A large scale distantly supervised challenge dataset for
  reading comprehension.
\newblock In {\em Proceedings of the 55th Annual Meeting of the Association for
  Computational Linguistics (Volume 1: Long Papers)}, volume~1,  1601--1611.

\bibitem[\protect\citeauthoryear{Mou \bgroup et al\mbox.\egroup
  }{2017}]{mou2017coupling}
Mou, L.; Lu, Z.; Li, H.; and Jin, Z.
\newblock 2017.
\newblock Coupling distributed and symbolic execution for natural language
  queries.
\newblock In {\em International Conference on Machine Learning},  2518--2526.

\bibitem[\protect\citeauthoryear{Pan \bgroup et al\mbox.\egroup
  }{2017}]{pan2017memen}
Pan, B.; Li, H.; Zhao, Z.; Cao, B.; Cai, D.; and He, X.
\newblock 2017.
\newblock Memen: Multi-layer embedding with memory networks for machine
  comprehension.
\newblock {\em arXiv preprint arXiv:1707.09098}.

\bibitem[\protect\citeauthoryear{Pennington, Socher, and
  Manning}{2014}]{pennington2014glove}
Pennington, J.; Socher, R.; and Manning, C.
\newblock 2014.
\newblock Glove: Global vectors for word representation.
\newblock In {\em Proceedings of the 2014 conference on empirical methods in
  natural language processing (EMNLP)},  1532--1543.

\bibitem[\protect\citeauthoryear{Rajpurkar \bgroup et al\mbox.\egroup
  }{2016}]{rajpurkar2016squad}
Rajpurkar, P.; Zhang, J.; Lopyrev, K.; and Liang, P.
\newblock 2016.
\newblock Squad: 100,000+ questions for machine comprehension of text.
\newblock In {\em Proceedings of the 2016 Conference on Empirical Methods in
  Natural Language Processing},  2383--2392.

\bibitem[\protect\citeauthoryear{Seo \bgroup et al\mbox.\egroup
  }{2016}]{seo2016bidirectional}
Seo, M.; Kembhavi, A.; Farhadi, A.; and Hajishirzi, H.
\newblock 2016.
\newblock Bidirectional attention flow for machine comprehension.
\newblock {\em arXiv preprint arXiv:1611.01603}.

\bibitem[\protect\citeauthoryear{Sutskever, Vinyals, and
  Le}{2014}]{sutskever2014sequence}
Sutskever, I.; Vinyals, O.; and Le, Q.~V.
\newblock 2014.
\newblock Sequence to sequence learning with neural networks.
\newblock In {\em Advances in neural information processing systems},
  3104--3112.

\bibitem[\protect\citeauthoryear{Tan \bgroup et al\mbox.\egroup
  }{2018}]{tan2017s}
Tan, C.; Wei, F.; Yang, N.; Lv, W.; and Zhou, M.
\newblock 2018.
\newblock S-net: From answer extraction to answer generation for machine
  reading comprehension.
\newblock In {\em In Proceedings of the 32th AAAI Conference on Artificial
  Intelligence}.

\bibitem[\protect\citeauthoryear{Vinyals, Fortunato, and
  Jaitly}{2015}]{vinyals2015pointer}
Vinyals, O.; Fortunato, M.; and Jaitly, N.
\newblock 2015.
\newblock Pointer networks.
\newblock In {\em Advances in Neural Information Processing Systems},
  2692--2700.

\bibitem[\protect\citeauthoryear{Wang and Jiang}{2016}]{wang2016machine}
Wang, S., and Jiang, J.
\newblock 2016.
\newblock Machine comprehension using match-lstm and answer pointer.
\newblock {\em arXiv preprint arXiv:1608.07905}.

\bibitem[\protect\citeauthoryear{Wang \bgroup et al\mbox.\egroup
  }{2017}]{wang2017gated}
Wang, W.; Yang, N.; Wei, F.; Chang, B.; and Zhou, M.
\newblock 2017.
\newblock Gated self-matching networks for reading comprehension and question
  answering.
\newblock In {\em Proceedings of the 55th Annual Meeting of the Association for
  Computational Linguistics (Volume 1: Long Papers)}, volume~1,  189--198.

\bibitem[\protect\citeauthoryear{Wang \bgroup et al\mbox.\egroup
  }{2018}]{wang2017r}
Wang, S.; Yu, M.; Guo, X.; Wang, Z.; Klinger, T.; Zhang, W.; Chang, S.;
  Tesauro, G.; Zhou, B.; and Jiang, J.
\newblock 2018.
\newblock R${}^3$: Reinforced reader-ranker for open-domain question answering.
\newblock In {\em In Proceedings of the 32th AAAI Conference on Artificial
  Intelligence}.

\bibitem[\protect\citeauthoryear{Weissenborn, Wiese, and
  Seiffe}{2017}]{weissenborn2017fastqa}
Weissenborn, D.; Wiese, G.; and Seiffe, L.
\newblock 2017.
\newblock Fastqa: A simple and efficient neural architecture for question
  answering.
\newblock {\em arXiv preprint arXiv:1703.04816}.

\bibitem[\protect\citeauthoryear{Xiong, Zhong, and
  Socher}{2016}]{xiong2016dynamic}
Xiong, C.; Zhong, V.; and Socher, R.
\newblock 2016.
\newblock Dynamic coattention networks for question answering.
\newblock {\em arXiv preprint arXiv:1611.01604}.

\bibitem[\protect\citeauthoryear{Zeiler}{2012}]{zeiler2012adadelta}
Zeiler, M.~D.
\newblock 2012.
\newblock Adadelta: an adaptive learning rate method.
\newblock {\em arXiv preprint arXiv:1212.5701}.

\end{thebibliography}
\bibliographystyle{aaai}

\end{document}